\documentclass[10pt,twocolumn,letterpaper]{IEEEtran}
\usepackage{amsmath,epsfig}
\usepackage{algorithm}
\usepackage{bm}
\usepackage{tzplot}
\usepackage[usenames,dvipsnames]{pstricks}
\usepackage{epsfig}
\usepackage{pst-grad} 
\usepackage{pst-plot} 
\usepackage{subfigure}
\usepackage[capitalize]{cleveref}

\usepackage{subfigure}
\usepackage[T1]{fontenc}
\usepackage{amsmath}
\usepackage{graphics}
\include{psfig}
\usepackage{epsfig}
\usepackage{array}
\usepackage{psfrag}
\usepackage{amssymb}
\usepackage{color}
\usepackage{pstricks,pst-node,pst-text,pst-3d,pst-plot}
\usepackage{cite}
\usepackage{ulem}
\definecolor{dgreen}{rgb}{0, 0.8, 0.1}

\begin{document}

\title{Modeling and Analysis of Spatial and Temporal Land Clutter Statistics in SAR Imaging Based on MSTAR Data \\
\author{\IEEEauthorblockN{Shahrokh Hamidi}\\
\IEEEauthorblockA{{Department of Electrical and Computer Engineering}, 
{University of Waterloo}\\
Waterloo, Ontario, Canada \\
shahrokh.hamidi@uwaterloo.ca}
}
}

\maketitle
\thispagestyle{empty}

\begin{tikzpicture}[remember picture, overlay]
      \node[font=\small] at ([yshift=-1cm]current page.north)  {This paper has been published  in the 2024 International Conference on Radar, Antenna, Microwave, Electronics, and Telecommunications (ICRAMET). \copyright IEEE};
\end{tikzpicture}

\begin{abstract}
The statistical analysis of land clutter for Synthetic Aperture Radar (SAR) imaging has become an increasingly important subject for research and investigation.  It is also absolutely necessary for designing robust algorithms capable of performing the task of target detection in the background clutter. Any attempt to extract the energy of the desired targets from the land clutter requires complete knowledge of the statistical properties of the background clutter. 
 
In this paper, the spatial as well as the temporal characteristics of the land clutter are studied. Since the data for each image has been collected based on a different aspect angle; therefore, the temporal analysis contains variation in the aspect angle. Consequently, the temporal analysis includes the characteristics of the radar cross section with respect to the aspect angle based on which the data has been collected. In order to perform the statistical analysis, several well-known and relevant distributions, namely, Weibull, Log-normal, Gamma, and Rayleigh are considered as prime candidates to model the land clutter.
The goodness-of-fit test is based on the Kullback-Leibler (KL) Divergence metric. The detailed analysis presented in this paper demonstrates that the Weibull distribution is a more accurate fit for the temporal-aspect-angle statistical analysis while the Rayleigh distribution models the spatial characteristics of the  background clutter with higher accuracy.  Finally, based on the aforementioned statistical analyses and by utilizing the Constant False Alarm Rate (CFAR) algorithm, we perform target detection in land clutter. 

The overall verification of the analysis is performed by exploiting the Moving and Stationary Target Acquisition and Recognition (MSTAR) data-set, which has been collected in spotlight mode at X-band, and the results are presented.    
\end{abstract}


\section{Introduction}
The target detection in clutter for  Synthetic Aperture Radar (SAR) images has been a subject of significant interest for decades. In order to extract the energy of the desired targets from the background clutter various techniques have been proposed in the literature which among them we can mention fixed threshold algorithm, Constant False Alarm Rate (CFAR) detection, Generalized Likelihood Ratio Test (GLRT), and Wavelet-based method \cite{fixed_threshold, Wavelet, SAR_Clutter, CFAR_Weibull, ground_clutter, clutter_threshold, CFAR_1, CFAR_2, CFAR_3, CFAR_4}. 
Any attempt to extract the energy of the desired targets based on a fixed threshold would most likely result in missing some features of  the desired targets while retaining the energy of the unwanted background clutter. The Wavelet-based methods also ignore the statistical properties of the clutter; therefore the result can still be considered as unsatisfactory. 

In this paper, the CFAR algorithm is utilized to design an adaptive threshold \cite{Skolnik, Mahafza}. The background clutter that we study in this work is land clutter. The CFAR-based threshold requires complete knowledge of the statistical properties of the background clutter. In order to perform this task, we study the spatial statistical properties of the background clutter thoroughly. The analysis is based on utilizing several well-known and relevant probability density functions (pdf), namely, Weibull, Log-normal, Gamma, and Rayleigh. The goodness-of-fit test is based on the Kullback-Leibler (KL) Divergence metric. 
We further perform temporal analysis for the land clutter. The experimental results, presented in this paper, is based on the Moving and Stationary Target Acquisition and Recognition (MSTAR) data-set \cite{MSTAR}, which has been collected in spotlight mode at X-band. The images have been created based on different aspect angles. Therefore, the temporal analysis includes the variation in aspect angle as well. 
Finally, we utilize the MSTAR data-set and present the results of the CFAR-based detection as well as the details of the entire process. 
The thorough analysis presented in this paper, demonstrates that the Weibull distribution can model the temporal-aspect-angle statistical properties of the background land clutter precisely while the Rayleigh distribution models the spatial characteristics of the background clutter with higher accuracy.  
Overall, both Weibull and Rayleigh distributions present themselves as perfect candidates to model the statistical properties of the land clutter spatially and temporally.

The paper has been organized as follows.  Section \ref{Statistical Modeling} discusses the statistical models that we consider in the paper. In Section \ref{CFAR-Based Detection}, we present the CFAR-based detection analysis. Finally, Section \ref{experimental results} has been dedicated to the experimental results based on the MSTAR data-set followed by the concluding remarks.

\section{Statistical Modeling}\label{Statistical Modeling}
In this section, we model the amplitude of the reflected signal from land clutter. The Weibull distribution is considered to be the main candidate for this task. Its pdf is expressed as \cite{Distribution} 
\begin{equation}
\label{Weibull}
p_W({x}| \alpha, \beta) =  \displaystyle \frac{\alpha}{\beta} {\left(\frac{x}{\beta}\right)}^{\alpha-1}e^{ \displaystyle - \left(\frac{x}{\beta}\right)^{\alpha}},  \;\;\; x \ge 0, \;\;\alpha, \beta > 0.
\end{equation}
The Log-normal distribution is the next option and its pdf is described as  \cite{Distribution}
\begin{equation}
\label{lognormal}
p_{\rm LN}({x}| \eta, \gamma) =  \displaystyle \frac{1}{x\eta \sqrt{2 \pi}} e^{ \displaystyle - \frac{(\log x -\gamma)^2}{2 \eta^2}},  \;\;\; x > 0, \;\; \eta > 0.
\end{equation}
The Gamma distribution with 2 degrees of freedom is the next best choice to model the land clutter. The pdf for the Gamma distribution is given as    \cite{Distribution}
\begin{eqnarray}
\label{Gamma}
p_G(x| a, b) = \frac{1}{b^a \Gamma(a)} x^{a-1}  e^{-x/b}, \;\;\; x>0, \;\; a, b > 0.
\end{eqnarray}
Finally, we consider the Rayleigh distribution as the last candidate to model the land clutter. The pdf for the Rayleigh distribution is expressed as   \cite{Distribution}
\begin{eqnarray}
\label{Rayleigh}
p_R(x| \sigma) = \frac{x}{\sigma^2} e^{\displaystyle -\frac{x^2}{2\sigma^2}}, \;\;\; x \ge 0, \;\; \sigma > 0.
\end{eqnarray}

In the next section, we elucidate the CFAR-based detection procedure. 
\section{CFAR-Based Detection}\label{CFAR-Based Detection}
The estimation of the adaptive threshold $T_a$ based on the definition of the probability of false alarm ($p_{\rm fa}$),  is given as \cite{Skolnik, Mahafza} 
\begin{equation}
\label{pfa}
p_{\rm fa} = p(X > T_a| H_0) = \int_{ T_a}^{\infty}  p(X| H_0) \; dX,
\end{equation}
in which $X$ describes the statistics of the cell under test and $T_a$ is the adaptive threshold. In addition, $H_0$ represents the null hypothesis. 

The KL divergence-based calculation will reveal that, among the proposed distributions, the Weibull and Rayleigh distributions are the best distributions to model the statistical properties of the background land clutter. Therefore, the calculation of the adaptive threshold is only performed for the Weibull and Rayleigh distributions. 
As a consequence, based on (\ref{pfa}), the estimation for the  adaptive threshold $T_{\rm aW}$ for the Weibull distribution, which has been given in (\ref{Weibull}), is given as
\begin{equation}
\label{thr_weibull}
T_{\rm aW} = \beta [ \log \left(\frac{1}{p_{\rm fa}}\right)]^{\displaystyle \frac{1}{\alpha}}.
\end{equation}
The estimated parameters of the Weibull distribution, given in (\ref{Weibull}), based on the maximum likelihood method are obtained as \cite{Weibull_ML}
\begin{eqnarray}
\label{params_weibull}
\hat{\beta} = \left(\frac{1}{N} \sum_{i = 1}^{N} x^{\hat{\alpha}}_i\right)^{\displaystyle \frac{1}{\hat{\alpha}}}, \nonumber \\
\hat{\alpha} = \displaystyle \frac{N}{\frac{1}{\hat{\beta}} \sum_{i = 1}^{N} x^{\hat{\alpha}}_i \log x_i - \sum_{i = 1}^{N} \log x_i},
\end{eqnarray}
in which $x_i$ represents the $i^{\rm th}$ realization of the random variable $x$.

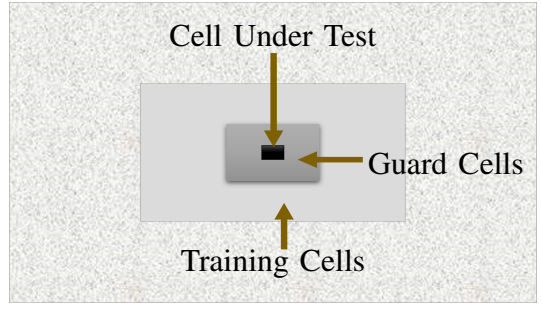
\begin{figure}
\centering
\begin{tikzpicture}[yshift=0.00001cm][font=\large]
\node(img1) {\includegraphics[height=4cm,width=7cm]{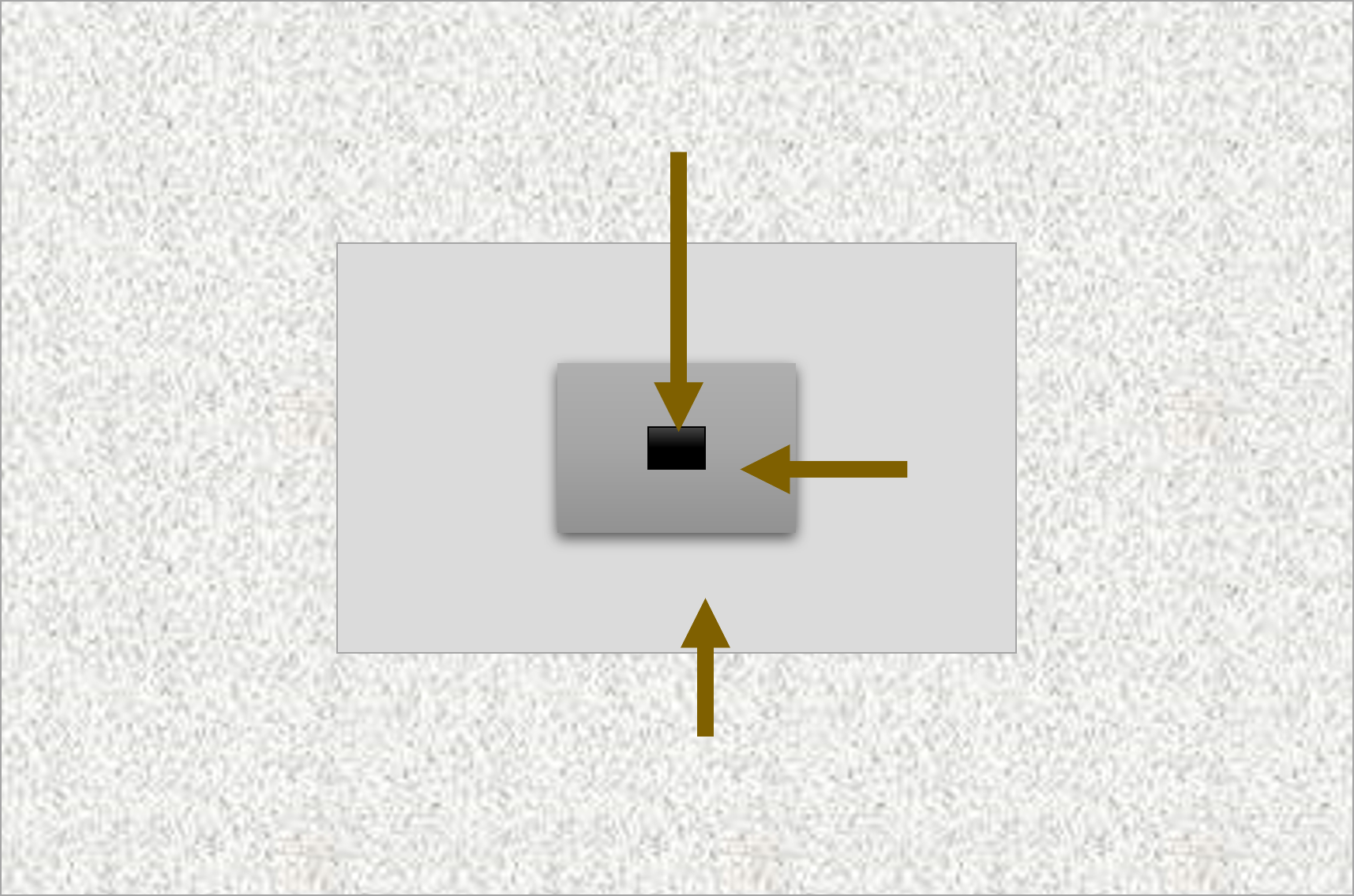}};
  \node[below=of img1, node distance=0cm, xshift=0cm, yshift=2cm,font=\color{black}] {{Training Cells}};
\node[below=of img1, node distance=0cm, xshift=2.3cm, yshift=3.3cm,font=\color{black}] {{Guard Cells}};
\node[below=of img1, node distance=0cm, xshift=0cm, yshift=5cm,font=\color{black}] {{Cell Under Test}};
\end{tikzpicture}
\caption{The structure of the 2D CFAR algorithm.}
\label{fig:CFAR}
\end{figure}

Furthermore, using (\ref{pfa}), the adaptive threshold $T_{\rm aR}$ for the Rayleigh distribution, which has been given in (\ref{Rayleigh}), is obtained as
\begin{equation}
\label{thr_rayleigh}
 T_{\rm aR} =  \sqrt{- 2\sigma^2 \log{(p_{\rm fa})}}.
\end{equation}
Moreover, the maximum likelihood estimate for the parameter of the Rayleigh distribution, given in (\ref{Rayleigh}), is described as
\begin{eqnarray}
\label{params_rayleigh}
\hat{\sigma} = \sqrt{ \frac{1}{2N}  \sum_{i = 1}^{N} x^2_i }. 
\end{eqnarray} 
In order to perform the task of target detection in land clutter, the 2D CFAR \cite{Skolnik, Mahafza} algorithm is implemented.  
The structure of the 2D CFAR algorithm has been displayed in Fig.~\ref{fig:CFAR}. 
Consequently, the decision making process for the cell under test is described as \cite{clutter_threshold, CFAR_Weibull}
\begin{equation}
\label{detection}
 X_{\rm CUT}  \mathop{\lessgtr}_{H_1}^{H_0}  \mu_c + \sigma_c  Q,
\end{equation}
where $X_{\rm CUT}$ is  the amplitude of the cell under test. Moreover, $\mu_c$ and $\sigma_c$ represent the sample mean and standard deviation which have been estimated from the clutter data of the local background. 
In addition, $Q = \frac{T_{\rm aW}}{\hat{\mu}}$ is the detector design parameter which defines the $p_{\rm fa}$ and is set empirically in which $ T_{\rm aW}$ is the adaptive threshold which has been described in (\ref{thr_weibull}) and $\hat{\mu}$ denotes the mean value estimated from the underlying clutter model. Furthermore, the $H_0$ and $H_1$ are the null and alternative hypotheses, respectively \cite{Kay, Mahafza}.    

The null hypothesis, $H_0$, represents the case in which the cell under test has been occupied by clutter while the alternative hypothesis, $H_1$, describes the case in which the cell contains the energy of the target.

\section{experimental results}\label{experimental results}

\subsection{Temporal Analysis}
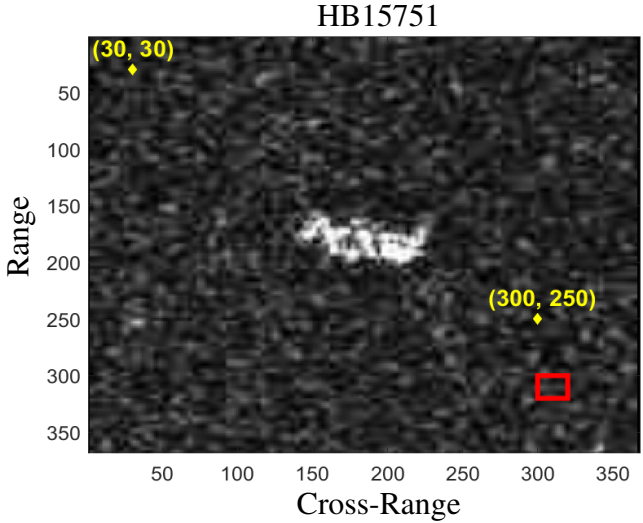
\begin{figure}
\centering
\begin{tikzpicture}[yshift=0.00001cm][font=\large]
  \node (img1)  {\includegraphics[height=6cm,width=8cm]{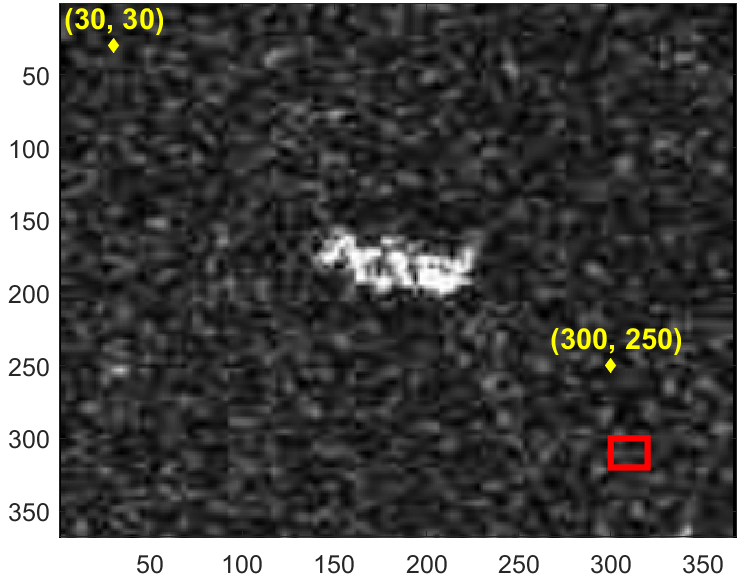}};
  \node[left=of img1, node distance=0cm, rotate = 90, xshift=1cm, yshift=-0.9cm,font=\color{black}] {{Range}};
  \node[below=of img1, node distance=0cm, xshift=0.5cm, yshift=1.2cm,font=\color{black}] {{Cross-Range}};
\node[below=of img1, node distance=0cm, xshift=0.5cm, yshift= 7.7cm,font=\color{black}] {{HB15751}};
\end{tikzpicture}
\caption{The original SAR image of BRDM-2 as well as two pixels, namely, $(30,30)^{\rm th}$ and $(300,250)^{\rm th}$, which are utilized to study the statistical properties of the background clutter. In addition, the data inside the red rectangle has been selected for further statistical analysis of the background land clutter.}
\label{fig:img_pixels_BRDM_2}
\end{figure}

\begin{figure}
\centering
\begin{tikzpicture}[yshift=0.00001cm][font=\large]
  \node (img1)  {\includegraphics[height=5.2cm,width=8cm]{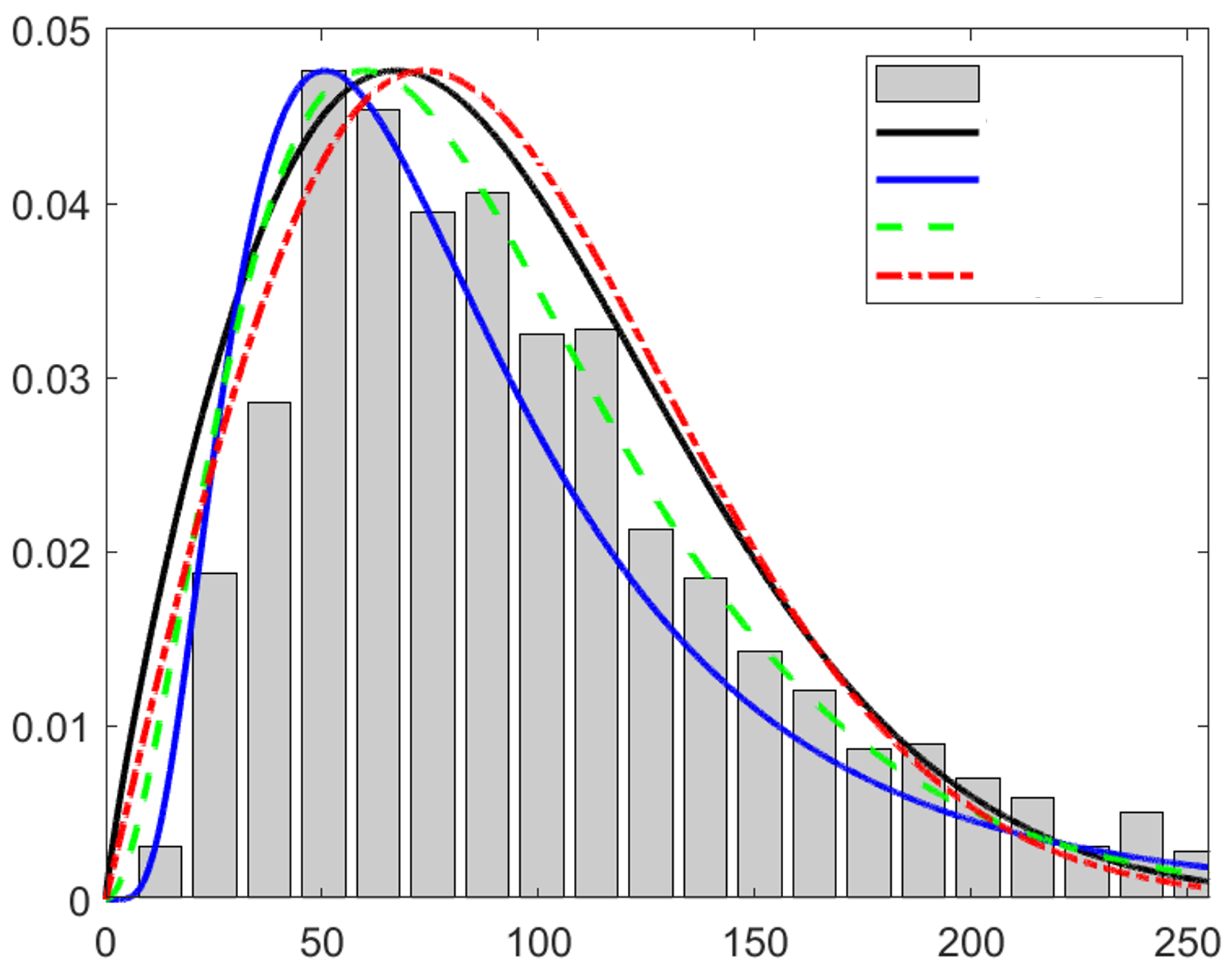}};
  \node[left=of img1, node distance=0cm, rotate = 90, xshift=1.2cm, yshift=-0.9cm,font=\color{black}] {{Frequency}};
  \node[below=of img1, node distance=0cm, xshift=0cm, yshift=1.2cm,font=\color{black}] {{Gray Level}};
 \node[below=of img1, node distance=0cm, xshift=3cm, yshift=6.1cm,font=\color{black}, font = {\tiny}] {{Histogram}};
 \node[below=of img1, node distance=0cm, xshift=3cm, yshift=5.85cm,font=\color{black}, font = {\tiny}] {{Weibull}};
 \node[below=of img1, node distance=0cm, xshift=3cm, yshift=5.6cm,font=\color{black}, font = {\tiny}] {{Log-normal}};
 \node[below=of img1, node distance=0cm, xshift=3cm, yshift=5.35cm,font=\color{black}, font = {\tiny}] {{Gamma}};
 \node[below=of img1, node distance=0cm, xshift=3cm, yshift=5.1cm,font=\color{black}, font = {\tiny}] {{Rayleigh}};
\end{tikzpicture}
\caption{The histogram as well as the estimated pdfs for the $(30,30)^{\rm th}$ pixel of the SAR image  which has been presented in Fig.~\ref{fig:img_pixels_BRDM_2}. }
\label{fig:hist_temporal_30_30}
\end{figure}
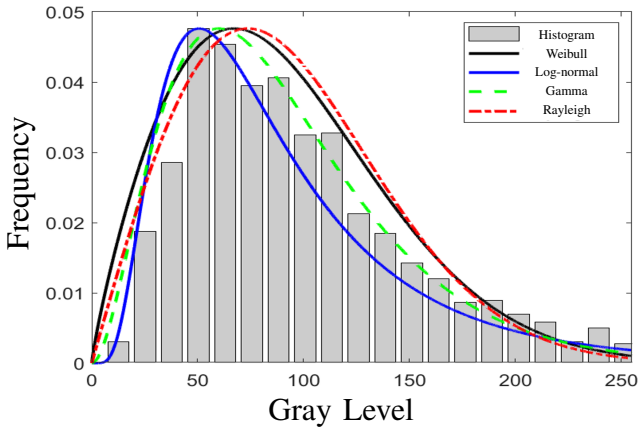

In this subsection, we present the temporal analysis for the land clutter. 
Fig.~\ref{fig:img_pixels_BRDM_2} shows the SAR image of BRDM-2. There are 1417 images for the BRDM-2 which have been created based on different aspect angles. Therefore, the temporal-aspect-angle statistical analysis for the background clutter is based on 1417 realizations. 
Fig.~\ref{fig:hist_temporal_30_30} illustrates the histogram as well as the estimated pdfs for the temporal data corresponding to the  $(30,30)^{\rm th}$ pixel of the SAR image which has been presented in Fig.~\ref{fig:img_pixels_BRDM_2}. The number of images which have been used to produce the result is 1417.  
In order to understand which distribution function is the most appropriate candidate to model the statistical properties of the land clutter, the KL distance is used
as a measure of fitness. For the estimated and empirical pdfs described as $p_e(x)$ and $p_d(x)$, respectively, the KL distance is expressed as 
\begin{equation}
\label{KL}
D_{\rm KL}{(p_d(x) || p_e(x))} = \int_{- \infty}^{\infty}   p_d(x) \log \left( \frac{p_d(x)}{p_e(x)} \right)   dx.
\end{equation}
The values for the KL distance for the given pdfs have been calculated based on (\ref{KL}) and have been presented in Table.\ref{Tab:Table_KL_1}.
\begin{table}
  \centering
  \caption{The values of the KL distance for the given  pdfs}\label{Tab:Table_KL_1}
  \begin{center}
    \begin{tabular}{| l | l | l |}

    \hline
    Distribution &  KL Distance  \\ \hline
    Weibull &  $29.6857$        \\ \hline
    Log-normal & $80.4051$     \\ \hline
    Gamma & $49.9830$     \\ \hline
    Rayleigh & $36.1063$     \\ \hline
    \hline
    \end{tabular}
\end{center}
\end{table}
From Table.\ref{Tab:Table_KL_1}, it is evident that, compared to the other probability distributions, the Weibull distribution, with the lowest KL distance, can model the statistical properties of the background clutter with higher accuracy. Furthermore,  the Rayleigh distribution is the second best fit which is followed by the Gamma and Log-normal distributions. 
 
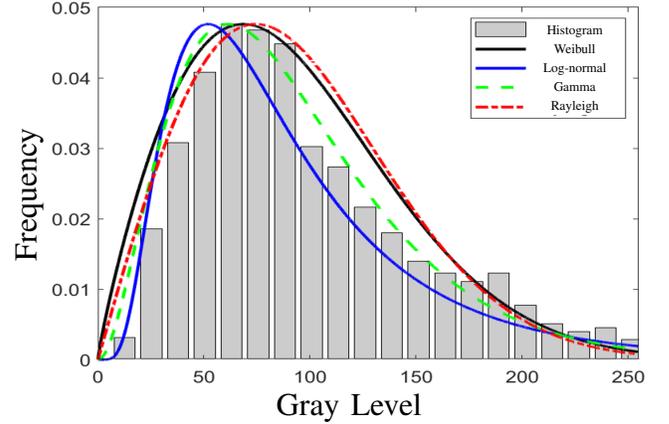
\begin{figure}
\centering
\begin{tikzpicture}[yshift=0.00001cm][font=\large]
  \node (img1)  {\includegraphics[height=5.2cm,width=8cm]{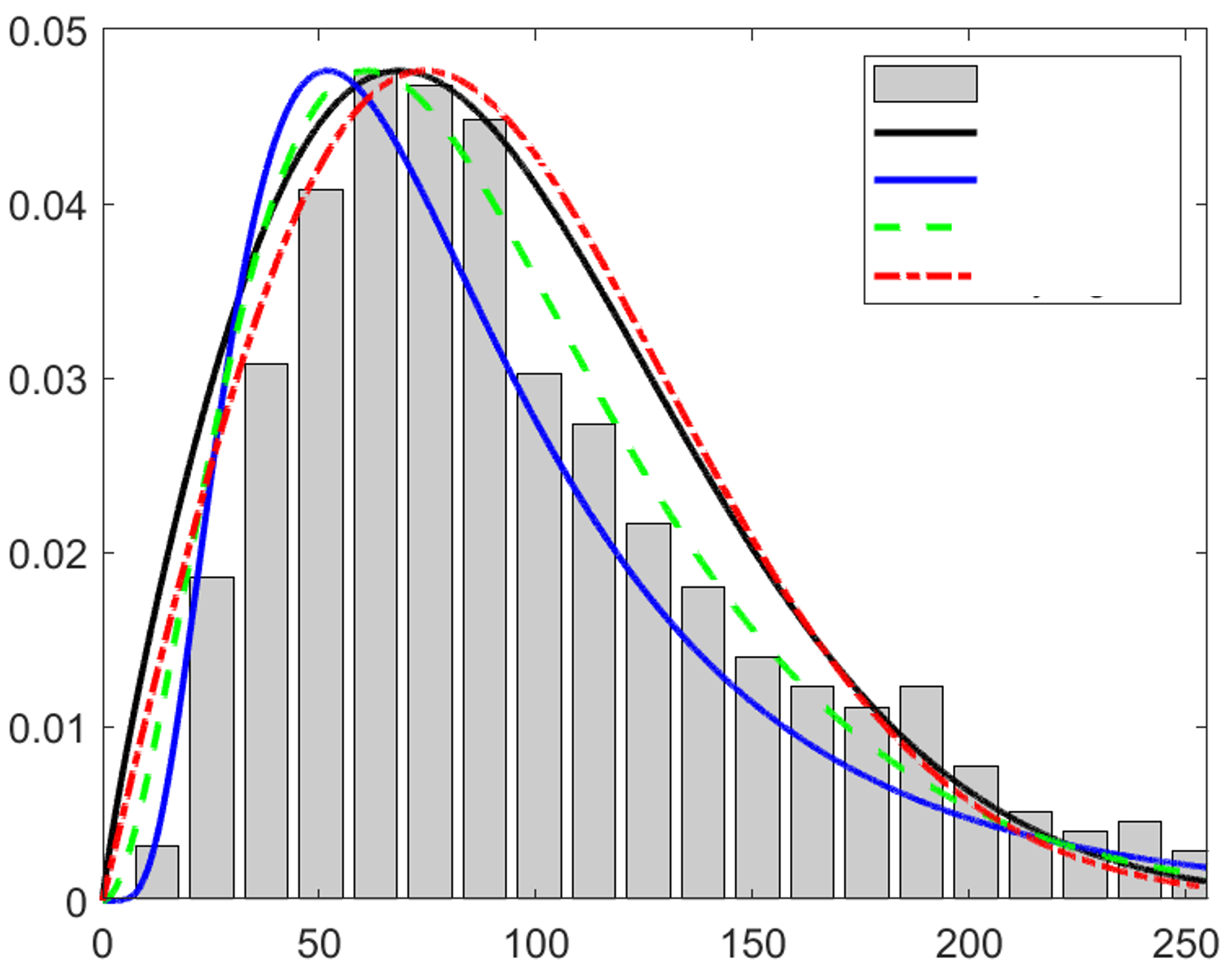}};
  \node[left=of img1, node distance=0cm, rotate = 90, xshift=1cm, yshift=-0.9cm,font=\color{black}] {{Frequency}};
  \node[below=of img1, node distance=0cm, xshift=0cm, yshift=1.2cm,font=\color{black}] {{Gray Level}};
 \node[below=of img1, node distance=0cm, xshift=3cm, yshift=6.1cm,font=\color{black}, font = {\tiny}] {{Histogram}};
 \node[below=of img1, node distance=0cm, xshift=3cm, yshift=5.85cm,font=\color{black}, font = {\tiny}] {{Weibull}};
 \node[below=of img1, node distance=0cm, xshift=3cm, yshift=5.6cm,font=\color{black}, font = {\tiny}] {{Log-normal}};
 \node[below=of img1, node distance=0cm, xshift=3cm, yshift=5.35cm,font=\color{black}, font = {\tiny}] {{Gamma}};
 \node[below=of img1, node distance=0cm, xshift=3cm, yshift=5.1cm,font=\color{black}, font = {\tiny}] {{Rayleigh}};
\end{tikzpicture}
\caption{The histogram as well as the estimated pdfs for the $(300,250)^{\rm th}$ pixel of the SAR image  which has been presented in Fig.~\ref{fig:img_pixels_BRDM_2}. }
\label{fig:hist_temporal_300_250}
\end{figure}

We have considered another pixel from a different part of the SAR image. Fig.~\ref{fig:hist_temporal_300_250} illustrates the histogram as well as the estimated pfds for the temporal data corresponding to the  $(300,250)^{\rm th}$ pixel of the SAR image which has been presented in Fig.~\ref{fig:img_pixels_BRDM_2}. The number of images which have been utilized to perform this analysis is 1417.  
The corresponding values for the KL distance based on the estimated pdfs have been calculated based on (\ref{KL}) and have been presented in Table.\ref{Tab:Table_KL_2}.
\begin{table}
  \centering
  \caption{The values of the KL distance for the given pdfs}\label{Tab:Table_KL_2}
  \begin{center}
    \begin{tabular}{| l | l | l |}

    \hline
    Distribution &  KL Distance  \\ \hline
    Weibull &  $27.7939$        \\ \hline
    Log-normal & $79.2342$     \\ \hline
    Gamma & $48.5611$  \\ \hline
    Rayleigh & $ 33.9155$     \\ \hline
    \hline
    \end{tabular}
\end{center}
\end{table}
From Table.\ref{Tab:Table_KL_2}, it is clear that compared to the other probability distributions, the Weibull distribution, with the lowest KL distance, can model the statistical properties of the land clutter with higher accuracy. And similar to the results that we achieved for the $(30,30)^{\rm th}$ pixel, the Rayleigh distribution has the second lowest KL distance followed by the Gamma and  Log-normal distributions.

\begin{figure}
\centering
\begin{tikzpicture}[yshift=0.00001cm][font=\large]
  \node (img1)  {\includegraphics[height=5cm,width=8cm]{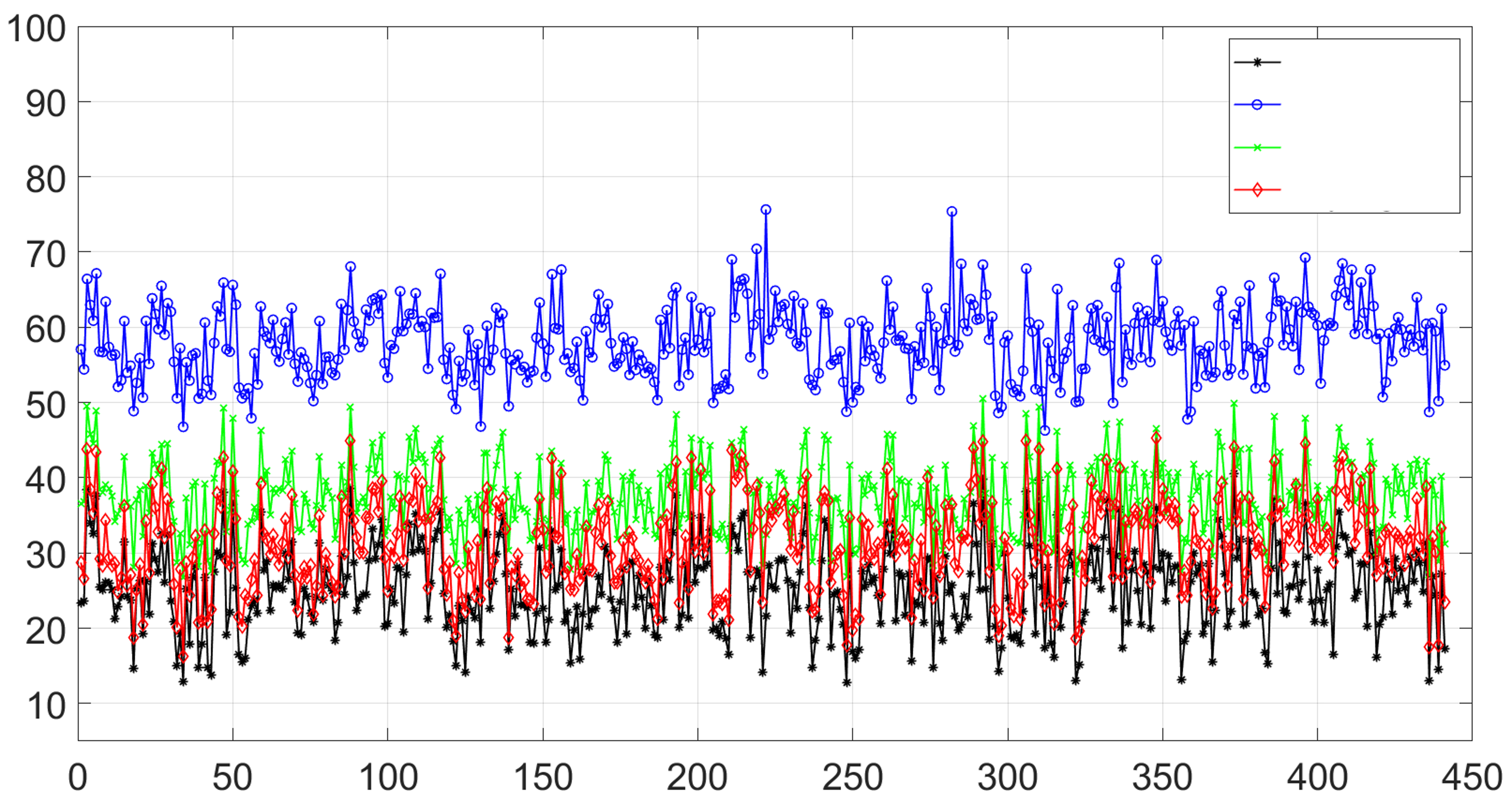}};
  \node[left=of img1, node distance=0cm, rotate = 90, xshift=1.3cm, yshift=-0.9cm,font=\color{black}] {{KL Distance}};
  \node[below=of img1, node distance=0cm, xshift=0cm, yshift=1.2cm,font=\color{black}] {{Pixels}};
 \node[below=of img1, node distance=0cm, xshift=3.2cm, yshift=5.97cm,font=\color{black}, font = {\tiny}] {{Weibull}};
 \node[below=of img1, node distance=0cm, xshift=3.25cm, yshift=5.73cm,font=\color{black}, font = {\tiny}] {{Log-normal}};
 \node[below=of img1, node distance=0cm, xshift=3.2cm, yshift=5.43cm,font=\color{black}, font = {\tiny}] {{Gamma}};
 \node[below=of img1, node distance=0cm, xshift=3.2cm, yshift=5.17cm,font=\color{black}, font = {\tiny}] {{Rayleigh}};
\end{tikzpicture}
\caption{The KL distance for different pixels related to the land clutter highlighted by the red rectangle in  Fig.~\ref{fig:img_pixels_BRDM_2} based on the proposed pdfs. As can be seen from the graph, the Weibull distribution has the lowest KL distance for almost all the pixels.  }
\label{fig:KLD}
\end{figure}
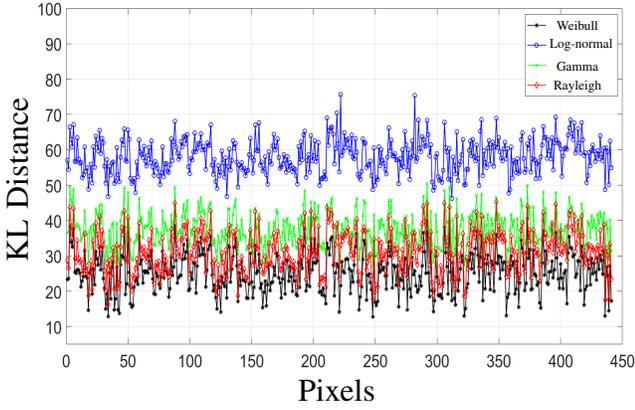

The next study that we conduct is for 441 pixels which have been highlighted by the red rectangle in Fig.~\ref{fig:img_pixels_BRDM_2}. The result of the analysis has been presented in Fig.~\ref{fig:KLD} which displays the KL distance for all the 441 pixels related to the background clutter and all the proposed pdfs. As can be seen from the graph, the Weibull distribution has the lowest KL distance for almost all the pixels followed by the Rayleigh distribution with the second lowest KL distance and Gamma and Log-normal distribution at the third and forth place, respectively. 

\subsection{Spatial Analysis}
\begin{figure}
\centering
\begin{tikzpicture}[yshift=0.00001cm][font=\large]
  \node (img1)  {\includegraphics[height=6cm,width=8cm]{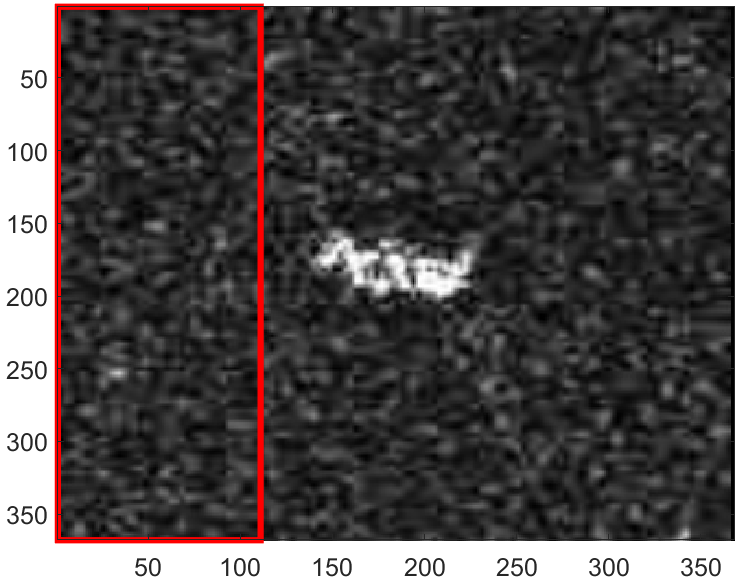}};
  \node[left=of img1, node distance=0cm, rotate = 90, xshift=1cm, yshift=-0.9cm,font=\color{black}] {{Range}};
  \node[below=of img1, node distance=0cm, xshift=0.5cm, yshift=1.2cm,font=\color{black}] {{Cross-Range}};
\node[below=of img1, node distance=0cm, xshift=0.5cm, yshift= 7.7cm,font=\color{black}] {{HB15751}};
\end{tikzpicture}
\caption{The original SAR image of BRDM-2 as well as the highlighted pixels in the red rectangle to be utilized for spatial statistical analysis.}
\label{fig:img_pixels_BRDM_2_spatial}
\end{figure}
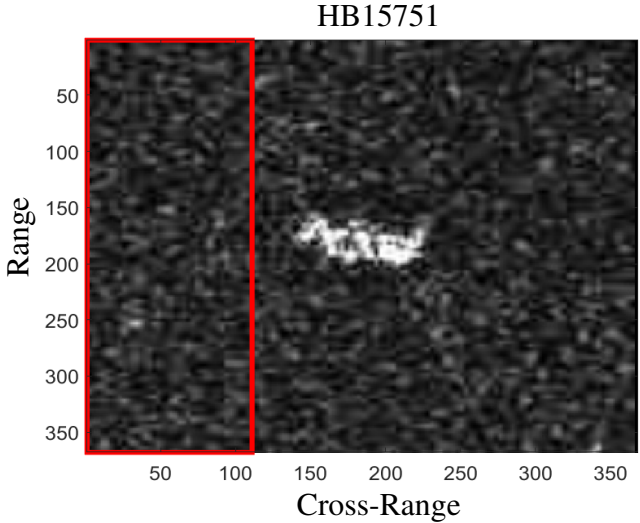

\begin{figure}
\centering
\begin{tikzpicture}[yshift=0.00001cm][font=\large]
  \node (img1)  {\includegraphics[height=5.5cm,width=8cm]{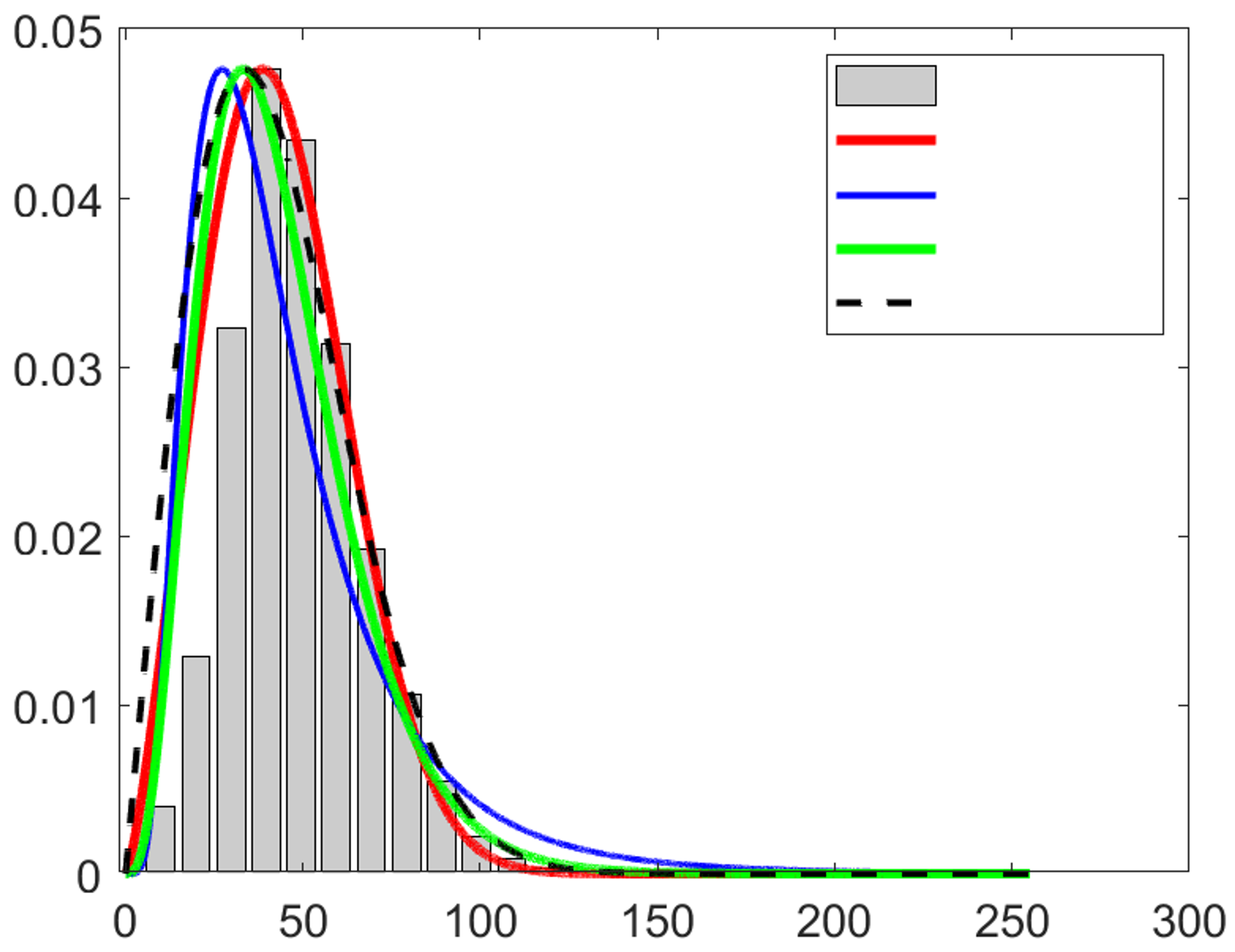}};
  \node[left=of img1, node distance=0cm, rotate = 90, xshift=1.2cm, yshift=-0.9cm,font=\color{black}] {{Frequency}};
  \node[below=of img1, node distance=0cm, xshift=0cm, yshift=1.2cm,font=\color{black}] {{Gray Level}};
 \node[below=of img1, node distance=0cm, xshift=2.8cm, yshift=6.43cm,font=\color{black}, font = {\small}] {{Histogram}};
 \node[below=of img1, node distance=0cm, xshift=2.85cm, yshift=6.08cm,font=\color{black}, font = {\small}] {{Weibull}};
 \node[below=of img1, node distance=0cm, xshift=2.88cm, yshift=5.78cm,font=\color{black}, font = {\small}] {{Log-normal}};
 \node[below=of img1, node distance=0cm, xshift=2.8cm, yshift=5.47cm,font=\color{black}, font = {\small}] {{Gamma}};
 \node[below=of img1, node distance=0cm, xshift=2.85cm, yshift=5.2cm,font=\color{black}, font = {\small}] {{Rayleigh}};
\end{tikzpicture}
\caption{The histogram as well as the estimated pdfs for the pixels inside the red rectangle in Fig.~\ref{fig:img_pixels_BRDM_2_spatial} to conduct the spatial statistical analysis. }
\label{fig:hist_spatial}
\end{figure}
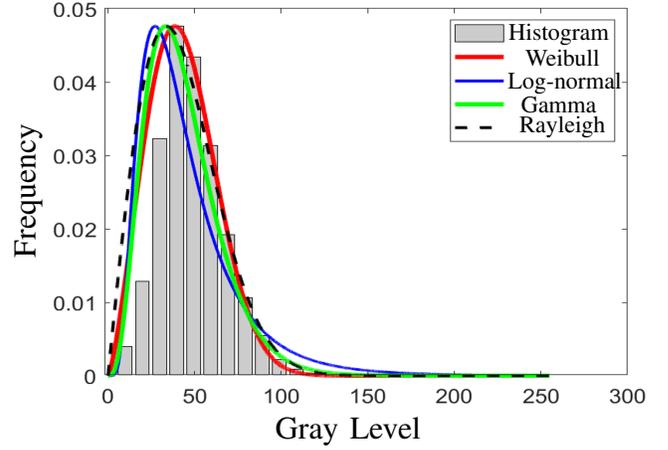

This subsection has been dedicated to the spatial analysis of the background land clutter. 
Fig.~\ref{fig:img_pixels_BRDM_2_spatial} shows the SAR image of BRDM-2 with the highlighted pixels in the red rectangle to be utilized for spatial statistical analysis.  
Fig.~\ref{fig:hist_spatial} illustrates the histogram as well as the estimated pdfs for the spatial data corresponding to the pixels inside the red rectangle in Fig.~\ref{fig:img_pixels_BRDM_2_spatial}.
The corresponding values for the KL distance based on the estimated pdfs have been calculated based on (\ref{KL}) and have been presented in Table.\ref{Tab:Table_KL_3}.
\begin{table}
  \centering
  \caption{The values of the KL distance for the given  pdfs}\label{Tab:Table_KL_3}
  \begin{center}
    \begin{tabular}{| l | l | l |}

    \hline
    Distribution &  KL Distance  \\ \hline
    Weibull &  $20.9582$        \\ \hline
    Log-normal & $31.6132$     \\ \hline
    Gamma & $25.8471$  \\ \hline
    Rayleigh & $16.0842$     \\ \hline
    \hline
    \end{tabular}
\end{center}
\end{table}
From Table.\ref{Tab:Table_KL_3}, it is evident that compared to the other probability distributions, the Rayleigh distribution, with the lowest KL distance, can model the statistical properties of the background land clutter with higher accuracy.

\begin{figure*}
\begin{tikzpicture}[yshift=0.00001cm][font=\small]
  \node (img1)  {\includegraphics[height=3.5cm,width=4.5cm]{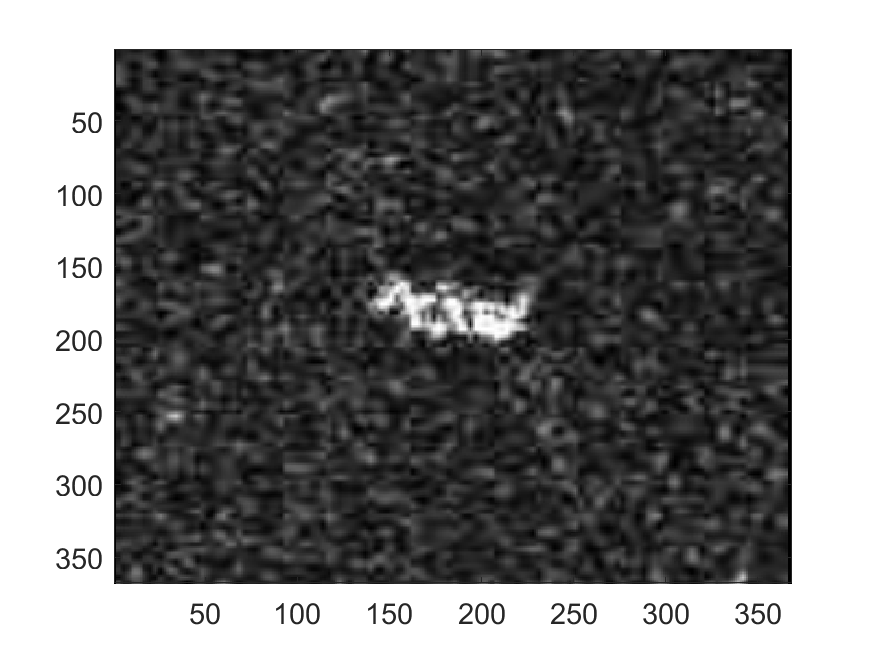}};
  \node[left=of img1, node distance=0cm, rotate = 90, xshift=0.7cm, yshift=-0.9cm,font=\color{black}] {{Images}};
\node[below=of img1, node distance=0cm, xshift=0cm, yshift= 4.9cm,font=\color{black}] {{HB15751}};
\hspace{4.2cm}
\node(img2) {\includegraphics[height=3.5cm,width=4.5cm]{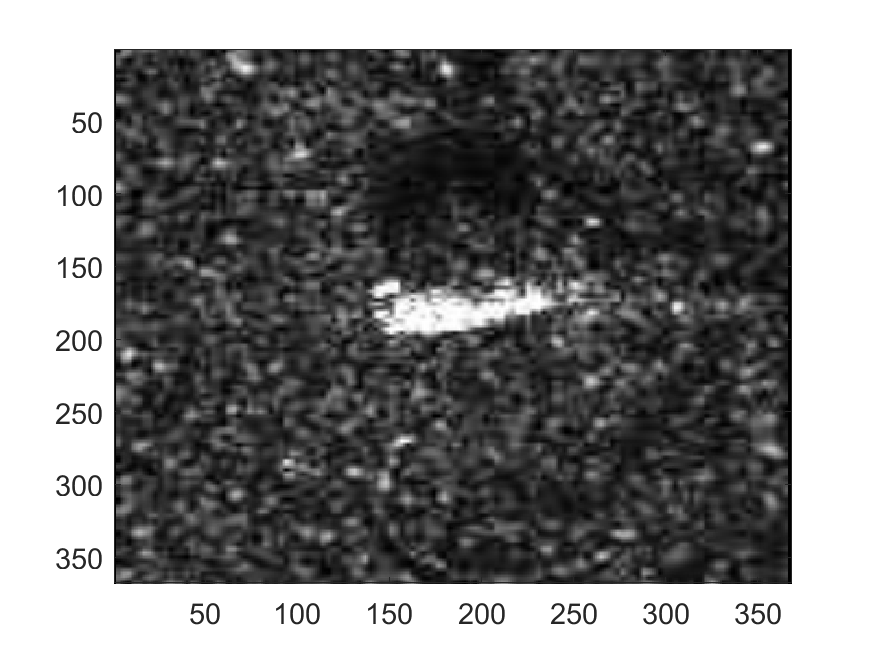}};
\node[below=of img1, node distance=0cm, xshift=0cm, yshift= 4.9cm,font=\color{black}] {{HB14951}};
\hspace{4.2cm}
\node(img3) {\includegraphics[height=3.5cm,width=4.5cm]{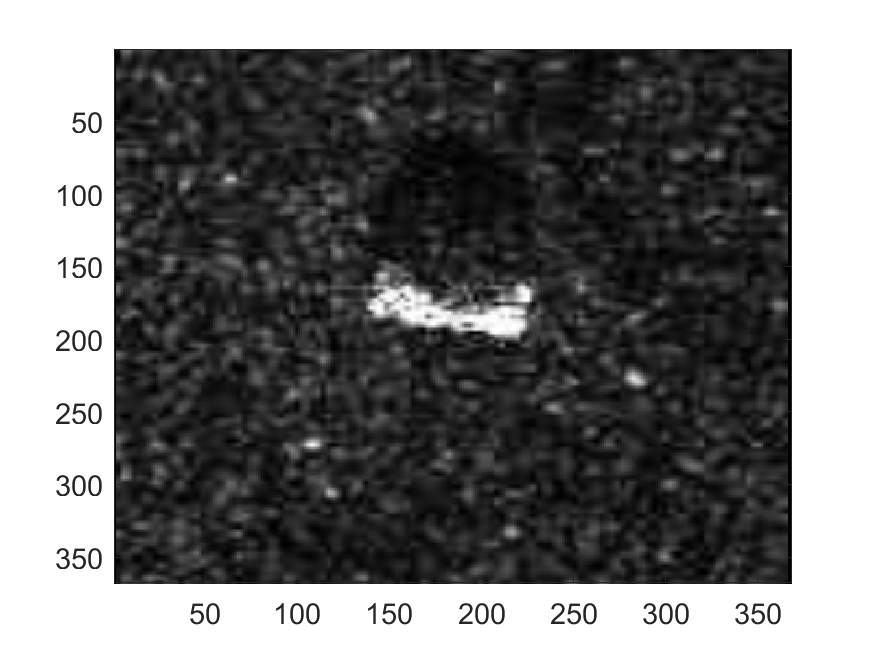}};
\node[below=of img1, node distance=0cm, xshift=0cm, yshift= 4.9cm,font=\color{black}] {{HB14955}};
\hspace{4.2cm}
\node(img4) {\includegraphics[height=3.5cm,width=4.5cm]{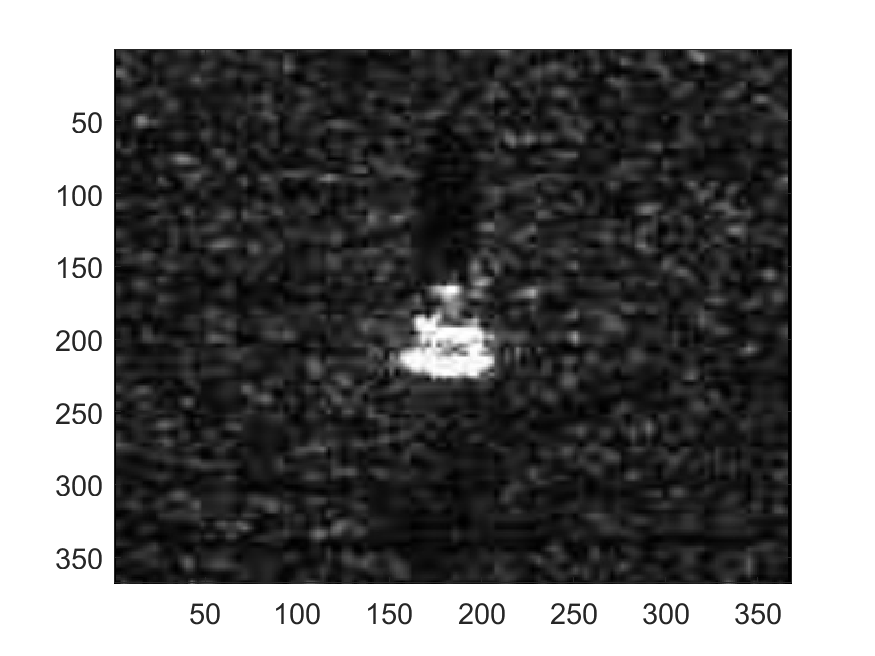}};
\node[below=of img1, node distance=0cm, xshift=0cm, yshift= 4.9cm,font=\color{black}] {{HB15041}};
\end{tikzpicture}

\begin{tikzpicture}[yshift=0.00001cm][font=\small]
  \node (img1)  {\includegraphics[height=3.5cm,width=4.5cm]{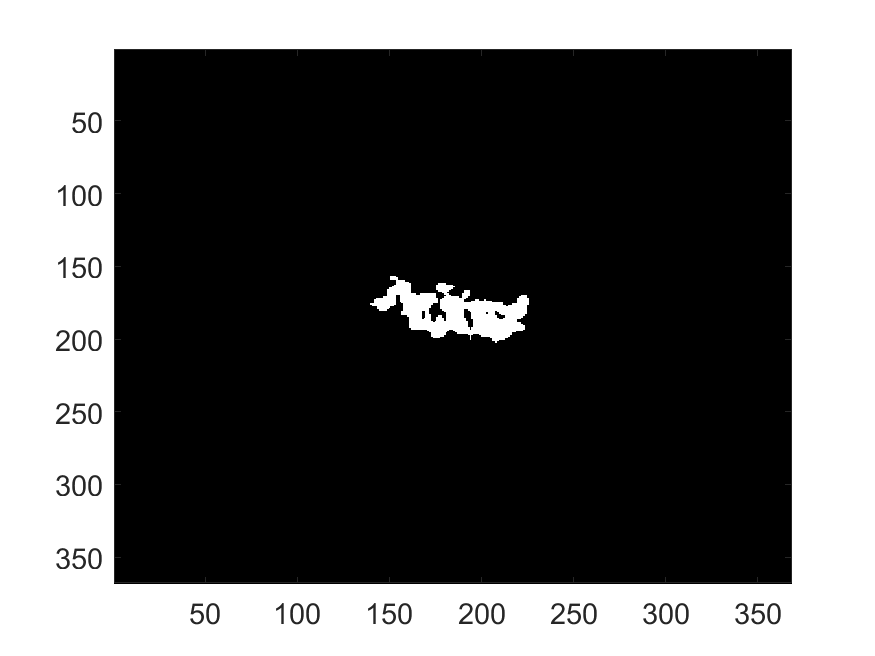}};
   \node[left=of img1, node distance=0cm, rotate = 90, xshift=1cm, yshift=-0.9cm,font=\color{black}] {{Weibull CFAR}};
\hspace{4.2cm}
\node(img2) {\includegraphics[height=3.5cm,width=4.5cm]{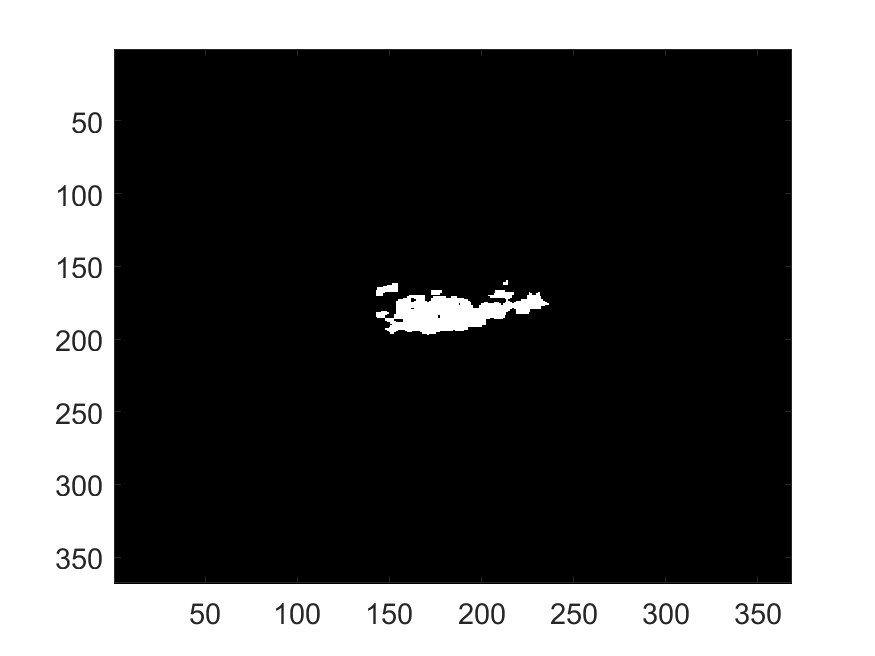}};
\hspace{4.2cm}
\node(img3) {\includegraphics[height=3.5cm,width=4.5cm]{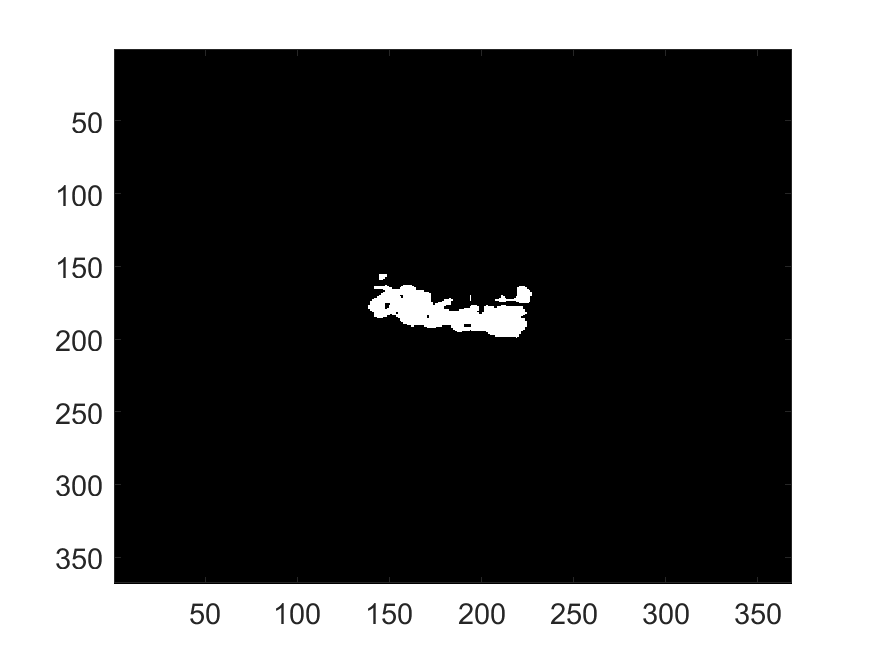}};
\hspace{4.2cm}
\node(img4) {\includegraphics[height=3.5cm,width=4.5cm]{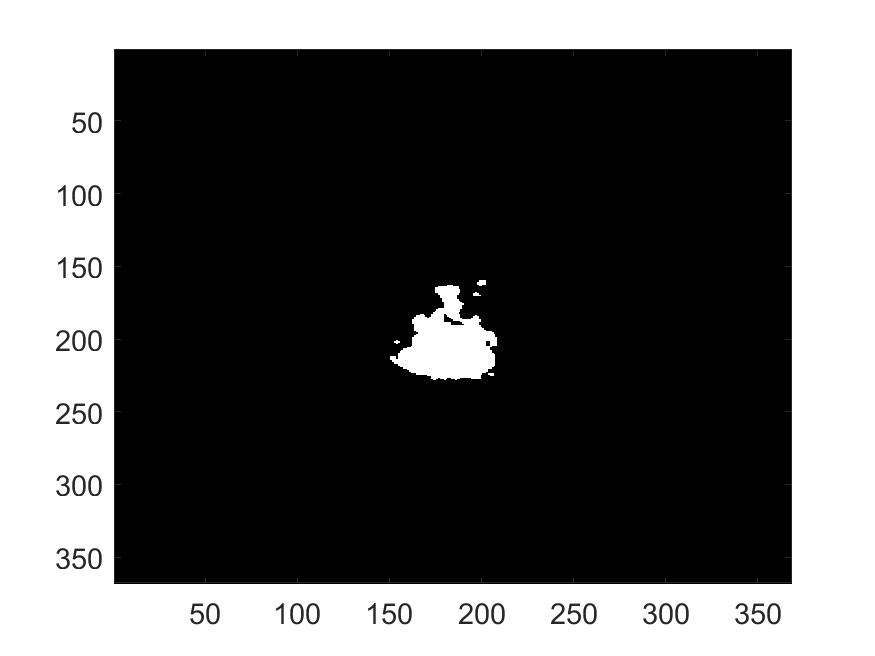}};
\end{tikzpicture}

\begin{tikzpicture}[yshift=0.00001cm][font=\small]
\centering
  \node (img1)  {\includegraphics[height=3.5cm,width=4.5cm]{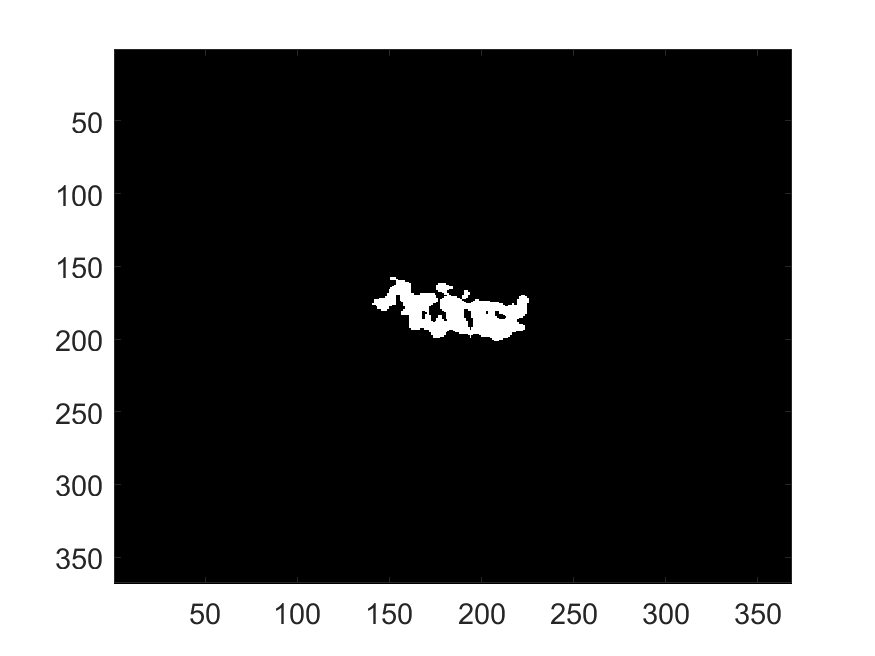}};
   \node[left=of img1, node distance=0cm, rotate = 90, xshift=1.2cm, yshift=-0.9cm,font=\color{black}] {{Rayleigh CFAR}};
\hspace{4.2cm}
\node(img2) {\includegraphics[height=3.5cm,width=4.5cm]{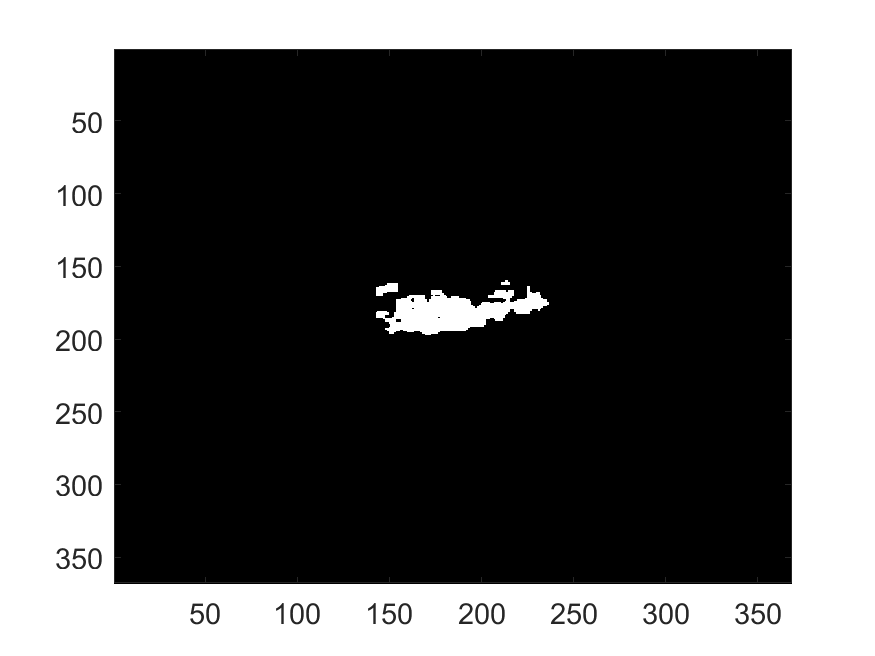}};
\hspace{4.2cm}
\node(img3) {\includegraphics[height=3.5cm,width=4.5cm]{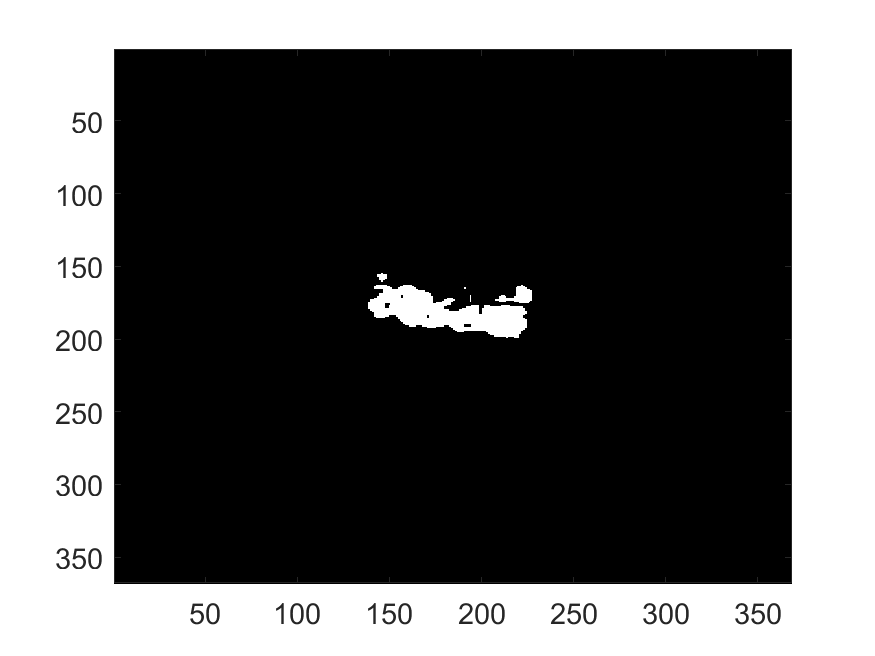}};
\hspace{4.2cm}
\node(img4) {\includegraphics[height=3.5cm,width=4.5cm]{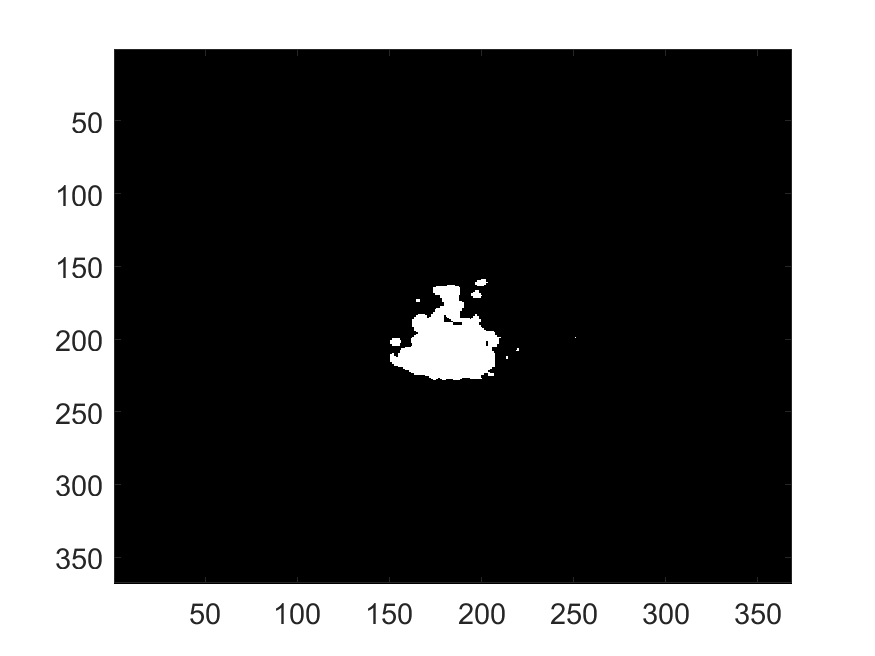}};
\end{tikzpicture}

\caption{The first row represents the MSTAR images. The second and the third rows illustrate the output of the CFAR algorithm for the Weibull- and Rayleigh-based detection. }
\label{fig:CFAR_res}
\end{figure*}
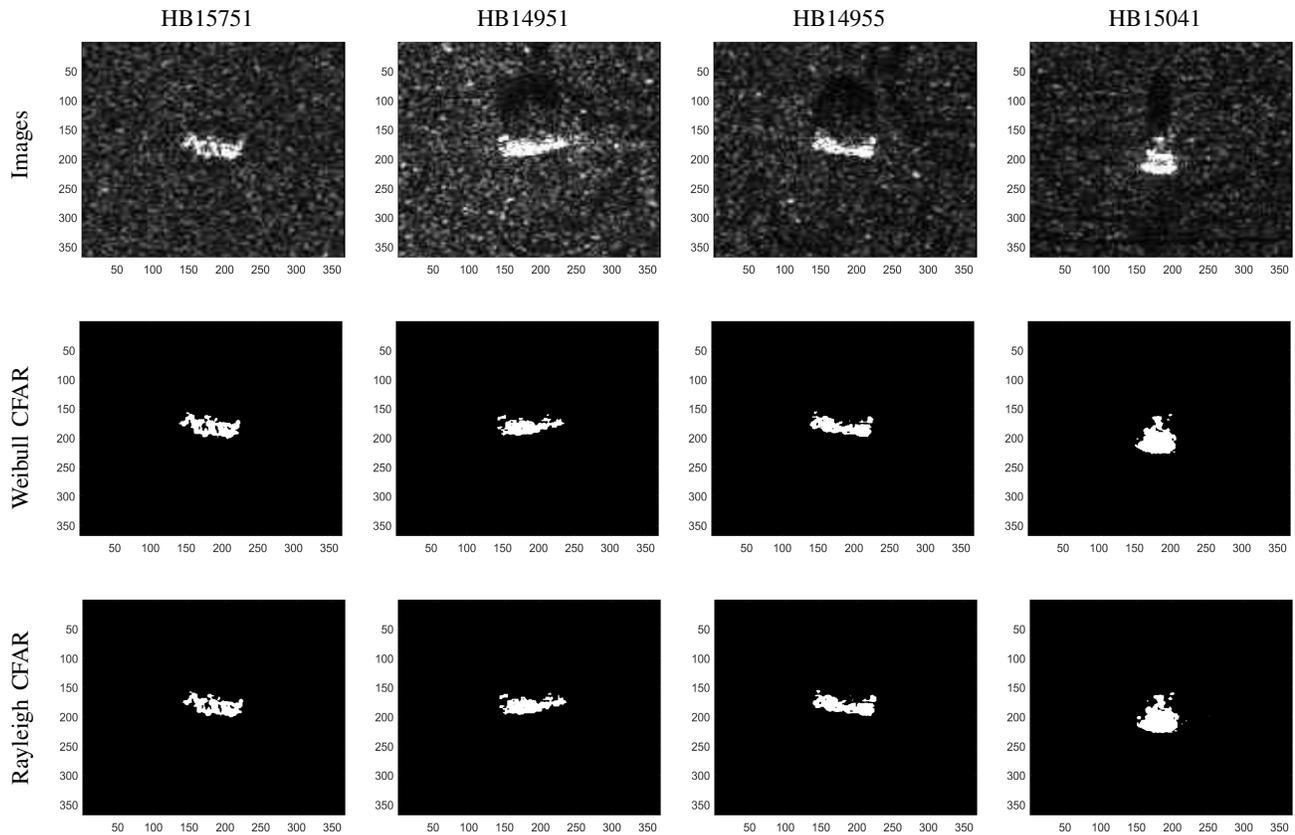

Based on the Weibull and Rayleigh distributions, we attempt to detect targets in background land clutter. The number of training cells for the upper, lower, left, and right wings of the cell under test have been set to 15. Moreover, we have selected the number of guard cells for the upper, lower, left, and right wings of the cell under test to be equal to 5.  We have further set the probability of false alarm as  $p_{\rm fa} = 10^{-6}$. The results have been displayed in Fig.~\ref{fig:CFAR_res}. The first row of Fig.~\ref{fig:CFAR_res} shows the images and the second and the third row represent the results of CFAR output for the Weibull and Rayleigh cases, respectively. 
The corresponding values for the $T_{\rm aW}$ and  $T_{\rm aR}$ based on (\ref{thr_weibull}) and (\ref{thr_rayleigh}) as well as $\mu_c$ and $\sigma_c$, the mean and standard deviation for the background land clutter, have been presented in Table.\ref{Tab:Table_KL_4}.
\begin{table}
  \centering
  \caption{The estimated values for the adaptive thresholds}\label{Tab:Table_KL_4}
  \begin{center}
    \begin{tabular}{| l | l | l | l | l |}

    \hline
             &           HB15751       &   HB14951      &  HB14955         &  HB15041       \\ \hline
    $T_{\rm aW}$ &     $164.5472$   &   $245.53$     &   $182.5098$    &  $150.3275$    \\ \hline
    $T_{\rm aR}$ &     $176.5383$    &   $227.996$   &  $165.4779$     &  $ 131.3287 $    \\ \hline
    $\mu_c$ &     $43.4732$    &   $54.6827$   &  $39.5334$     &  $31.2317$    \\ \hline
    $\sigma_c$ &     $19.1296$    &   $27.7925$   &  $20.4734$     &  $16.5223$    \\ \hline
    \hline
    \end{tabular}
\end{center}
\end{table} 

From Fig.~\ref{fig:CFAR_res}, it is apparent that both the Weibull and Rayleigh distributions provide excellent results when it comes to detecting  targets buried in the background land clutter.  

\section{conclusion}
Spatial and temporal statistical analyses of the land clutter for SAR images were presented. Accurate statistical models for each case were studied and the acquired information was used to design an adaptive threshold based on CFAR algorithm. At the end, the effectiveness of the proposed approach was assessed by using the MSTAR data-set and the results were presented and discussed in detail.

%

\begin{IEEEbiography}[{\includegraphics[width=1in,height=1.25in,clip,keepaspectratio]{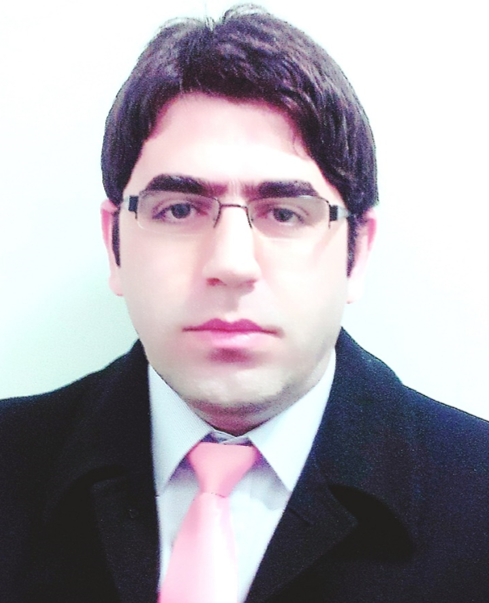}}]{Shahrokh Hamidi} was born in 1983, in Iran. He received his B.Sc., M.Sc., and Ph.D. degrees all in Electrical and Computer Engineering. He is with the Faculty of Electrical and Computer Engineering at the University of Waterloo, Waterloo, Ontario, Canada. His current research areas include statistical signal processing, mmWave imaging, Terahertz imaging, image processing, system design,  multi-target tracking, wireless communication, machine learning, optimization, and array processing.
\end{IEEEbiography}

\end{document}